# ORTHOSEG: A DEEP MULTIMODAL CONVOLUTIONAL NEURAL NETWORK ARCHITECTURE FOR SEMANTIC SEGMENTATION OF ORTHOIMAGERY

Pankaj Bodani[1, *], Kumar Shreshtha[2], Shashikant Sharma[1]

[1] Space Applications Centre, ISRO, Ahmedabad, Gujarat, India - (pankajb, sasharma)@sac.isro.gov.in
[2] Birla Institute of Technology and Science, Pilani, Rajasthan, India - f2015873@pilani.bits-pilani.ac.in

**Commission V, SS: Emerging Trends in Remote Sensing**

**KEY WORDS:** Deep Learning, Supervised Image Segmentation, Residual Networks

**ABSTRACT:**

This paper addresses the task of semantic segmentation of orthoimagery using multimodal data e.g. optical RGB, infrared and digital surface model. We propose a deep convolutional neural network architecture termed OrthoSeg for semantic segmentation using multimodal, orthorectified and coregistered data. We also propose a training procedure for supervised training of OrthoSeg. The training procedure complements the inherent architectural characteristics of OrthoSeg for preventing complex co-adaptations of learned features, which may arise due to probable high dimensionality and spatial correlation in multimodal and/or multispectral coregistered data. OrthoSeg consists of parallel encoding networks for independent encoding of multimodal feature maps and a decoder designed for efficiently fusing independently encoded multimodal feature maps. A softmax layer at the end of the network uses the features generated by the decoder for pixel-wise classification. The decoder fuses feature maps from the parallel encoders locally as well as contextually at multiple scales to generate per-pixel feature maps for final pixel-wise classification resulting in segmented output. We experimentally show the merits of OrthoSeg by demonstrating state-of-the-art accuracy on the ISPRS Potsdam 2D Semantic Segmentation dataset. Adaptability is one of the key motivations behind OrthoSeg so that it serves as a useful architectural option for a wide range of problems involving the task of semantic segmentation of coregistered multimodal and/or multispectral imagery. Hence, OrthoSeg is designed to enable independent scaling of parallel encoder networks and decoder network to better match application requirements, such as the number of input channels, the effective field-of-view, and model capacity.

## 1. INTRODUCTION

The phenomenal success of deep learning in image classification has inspired its use for the task of semantic image segmentation. Several supervised deep learning neural network architectures that perform segmentation by classifying every pixel into one of the target classes have met with great success in a number of applications (Chen et al., 2018, Badrinarayanan et al., 2017, Ronneberger et al., 2015). These architectures are usually derived from architectures originally designed for classification. However, this adaptation of deep learning based classification models to semantic segmentation is not trivial and possesses its own set of challenges such as coarse segmentation boundaries, localization errors, computational and memory requirements and requirement of very large training sets. Many novel techniques and modifications to state-of-the-art classification networks have been proposed to overcome these problems and to make the network suitable for pixel-wise classification (Chen et al., 2018, Badrinarayanan et al., 2017, Ronneberger et al., 2015). Semantic segmentation using classification models as pre-trained feature extractors has also been successfully demonstrated (Piramanayagam et al., 2018) (Liu et al., 2017, Sherrah, 2016). Semantic segmentation is particularly challenging when it involves multimodal data for which it may be difficult to obtain large training datasets or suitable pre-trained feature extractors. Multimodal data is very commonly used in remote sensing applications. This can be in the form of optical images in the visible spectrum, digital surface models captured using LIDAR, SAR backscatter, and multispectral imagery with non-visual channels. Therefore, design and implementation of efficient network architectures that can effectively use multimodal training datasets of limited size is an active area of research and development (Xiong et al., 2016, Kaiser et al., 2017). In this paper, we present a neural network architecture we call OrthoSeg for semantic segmentation using multimodal, orthorectified and co-registered imagery. Adaptability to varying application requirements and multimodal input domains, computational efficiency and ability to generalize well despite limited training data are key architectural goals of OrthoSeg.

## 2. LITERATURE REVIEW

Most of the recent architectures proposed for semantic segmentation are based on fully convolutional network (FCN) (Shelhamer et al., 2017) design approach wherein we get rid of the fully connected layers completely and the network is made entirely out of convolutional layers. It allows to somewhat preserve the spatial correlation among adjacent image pixels, something which is essential to semantic segmentation and is inevitably lost with the presence of fully connected layers. A major challenge in adapting the classification networks for the task of segmentation is the fact that while the deeper layers learn more rich higher order features, the spatial information in the feature maps is significantly lost because of strided convolutions and max-pooling operations. Getting rid of max-pooling and/or strided convolution is challenging as they not only decrease the computation and memory requirements but more importantly help the network learn richer features by providing a greater field-of-view, allowing it to respect the contextual information and prevent it from making locally optimal decisions, something which is essential to the performance of any good semantic segmentation algorithm. Several techniques like hypercolumn methods (Hariharan et al., 2015), skip connections from initial to final layers (Badrinarayanan et al., 2017, Lin et al., 2016, Ronneberger et al., 2015) and dilated/atrous convolutions (Chen et al., 2018) have been suggested to tackle this trade-off between learning richer representations and preserving spatial features.

* Corresponding author

As mentioned above, the observation that the initial layers of a deep convolutional network have greater amount of spatial information with lower order feature representations while the deeper layers learn much more sophisticated features but loose the spatial information has been a key motivation and insight in the design and success of all the state-of-art segmentation networks. All these networks essentially compensate for this loss of spatial information by passing this information from the initial layers to the later layers, mostly in form of skip connections. This allows the network to learn a combined representation of large, coarse higher order features and small, fine, lower order spatial features.

While hypercolumn methods and skip connections aim on regaining the spatial information lost due to max-pooling/strided convolutions by using the information preserved in initial layers, atrous convolution presents a technique to get rid of down-sampling itself. An atrous/dilated convolution is a convolution operation in which the filter weights have been spaced out, being separated from their adjacent weights by a fixed amount that is determined by the dilation rate. This provides an effectively larger filter and thus an increased receptive field while keeping the number of parameters constant. It allows the architecture to have an increased field-of-view without the need for down-sampling. The idea is to double the dilation rate of all subsequent convolutions for every deleted max-pooling layer, this way preserving the field-of-view that would have resulted as a consequence of max-pooling while also keeping the dimensionality of the feature maps intact and thus preserving spatial information. However, in practice, max-pooling provides more than just an increased receptive field and is found to be an essential component for network's performance. Hence, for this reason and for computational/memory efficiency, in the networks that make use of dilated convolutions for this purpose, max-pooling is replaced with dilated convolutions only for the later layers, once the image has been down-sampled by at least a factor of 8.

Simple encoder-decoder based architectures that try to construct segmented output from the highly spatially compressed feature representations by learning deconvolution layers from the higher order feature map, with the help of spatial information from earlier layers propagated through skip connections to merge these dense features with lower level spatial features, have been found to work really well in a wide variety of applications (Badrinarayanan et al., 2017, Ronneberger et al., 2015). OrthoSeg is also based on similar encoder-decoder principle but with many significant differences, as discussed later, in order to adapt it for the challenges faced in semantic segmentation of complex mutimodal high resolution imagery which arise due to a large field of-view requirement combined with involvement of multimodal features which result in high dimensionality.

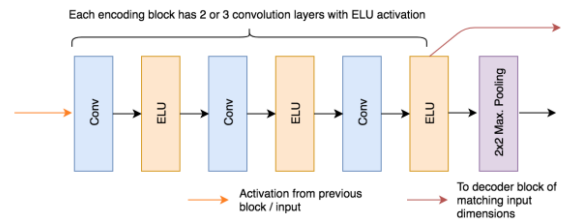

Figure 1. Encoder Block

## 3. ARCHITECTURE

OrthoSeg has multiple parallel encoder networks and a common decoder network which is followed by a softmax layer for final pixel-wise classification. Each of the encoder networks is topologically similar to the convolutional part of VGG-16 (Simonyan and Zisserman, 2014). The number of encoding blocks in the encoder can be tuned as per application specific field-of-view requirements. The layers in each encoding block are exactly same as VGG-16 as illustrated in Figure 1, except for the fact that we use ELU (Clevert et al., 2015) activation for all trainable blocks. The primary and the auxiliary encoders are identical in structure except for the number of filters in each convolutional layer which is a tuneable hyperparameter, allowing for relative scaling of capacity of each encoder depending on relative importance and information content of multimodal inputs. OrthoSeg also comprises a spatial correlation correction block (SCCB) which corrects decision activations from the decoder by learning spatial correlation probabilities of different classes using average pooling and dilated convolutions followed by two 1x1 convolutional layers. The purpose of SCCB is to learn to correct the residual error in class-wise activations generated by the decoder based on agreement with class-wise activations in close spatial neighbourhood. The design of SCCB is illustrated in Figure 2.

The decoder network has a block corresponding to each block of the encoder network. The design of the decoder block is illustrated in Figure 3. Every decoder block up-samples the input feature activations and concatenates it with outputs of the pre-pooling convolution layers of the corresponding blocks in the two parallel encoder networks. Decoder block also receives scaled network input and generates an additional feature map by subtracting 5x5 pixel moving average from the input. This is then concatenated with up-sampled feature activations and followed by two atrous convolution layers with dilation rate of two with ELU activation. The feature activations are fed into a convolution layer with 1x1 kernel which generates class-wise decision activations which are added to input class-wise decision activations in a residual manner.

The output of the final decoder block is per-pixel feature activations and class wise decision activations which are fed directly to SCCB or via multiple additional residual learning blocks illustrated in Figure 4. SCCB is followed by softmax activation. The result is one-hot encoded class probabilities for final pixel-wise classification. The activation connectivity of the decoder, additional residual blocks and SCCB is illustrated in Figure 5.

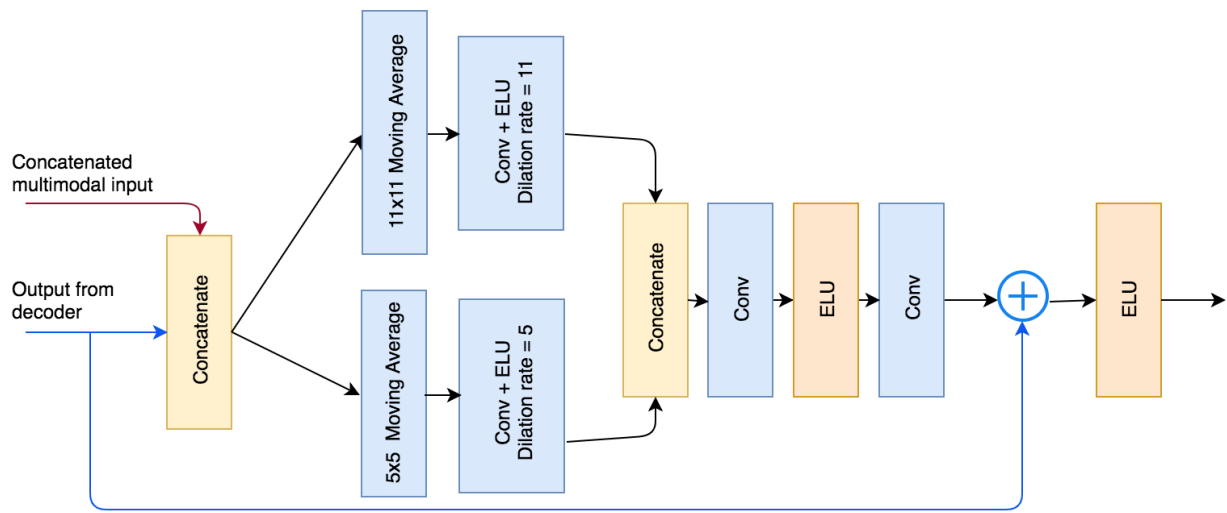

Figure 2. Spatial correlation correction block

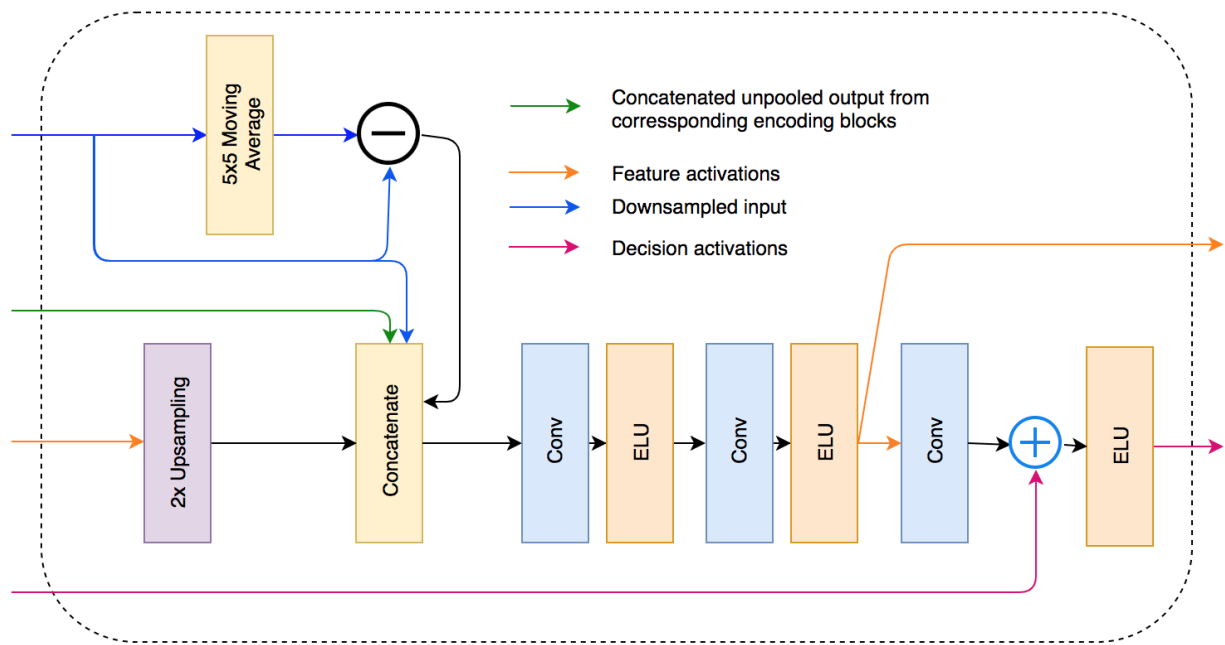

Figure 3. Decoder Block

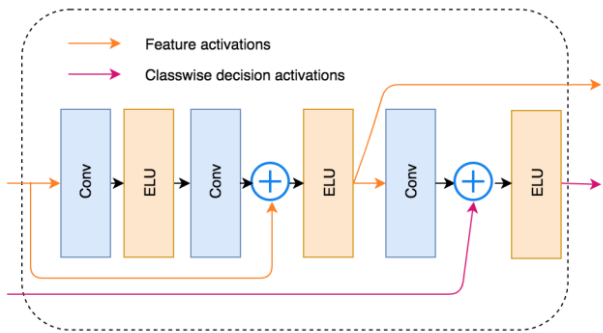

Figure 4. Additional Residual Block

Although our architecture has some similarity with the existing networks for semantic segmentation in the 'auto-encoders with shortcuts' design paradigm (Badrinarayanan et al., 2017, Ronneberger et al., 2015) it has several key differences. The most important difference is that all decoders contribute directly to final class probabilities through direct connection to softmax layer via SCCB. Second, instead of using decoder blocks with tapering number of filters, we use constant filter depth in all decoding blocks and add residual connection for each block. Additionally, each decoder block is also directly fed the input of the network down-scaled to the input dimensions of the decoder. Finally, the gradients do not flow from all paths in the network. The flow of gradients during backpropagation is only allowed through skip connections in the decoder block and not to previous decoder block. SCCB gradients do not flow back to the decoder network. The motivation behind this design is to effectively create an architecture that mimics an ensemble of decoders which independently contribute to decision at multiple spatial scales while retaining the ability of coarse upstream decoders to communicate feature activations to finer downstream decoders.

An instance of OrthoSeg that we used for the benchmark test is illustrated in Figure 6. The two parallel encoders take as input IR-Red-Green (IRRG) and Blue-NDVI-DSM channels respectively and both have seven encoding blocks. The outputs from the final layers of the two parallel encoders are concatenated and fed to the decoder. We call the encoder network with IR-Red-Green inputs as the primary encoder and the encoder network with Blue-NDVI-DSM input as the auxiliary encoder. The first five blocks of the primary encoder and first three blocks of the secondary encoder are initialized with VGG-16 pre-trained weights on the ImageNet (Deng et al., 2009) dataset for image classification (Chollet et al., 2015). The weights on of the rest of the network were randomly initialized using technique described in (He et al., 2015). The number of convolutional filters that we used in each block for the Potsdam benchmark challenge is given in Table 1. All convolutional layers generating decision activations have one filter for each class. We used a single additional residual block in our network.

### 3.1 Dataset

We use the ISPRS Potsdam 2D semantic labelling challenge dataset to demonstrate the performance of OrthoSeg. The dataset consists of 38 patches of 6000x6000 pixels extracted from a larger true orthophoto (TOP) mosaic of the city of Potsdam, Germany. The images consist of four channels i.e. red, green, blue and infrared. Additionally, gridded digital surface model in form of single band grayscale image consisting of raw height values in metres are is provided for each of the 38 patches. Of the 38 patches, 24 are training patches and 14 are test patches. Each pixel in the images is to be classified into one of the following six classes:

1. Impervious Surfaces – White
2. Building – Dark Blue
3. Low Vegetation – Light Blue
4. Tree – Green
5. Car – Yellow
6. Clutter/Background – Red

| Block/Layer | Primary Encoder | Auxiliary Encoder |
|---|---|---|
| Block 1 | 64 | 64 |
| Block 2 | 128 | 128 |
| Block 3 | 256 | 256 |
| Block 4 | 512 | 256 |
| Block 5 | 512 | 256 |
| Block 6 | 512 | 256 |
| Block 7 | 512 | 256 |
| Decoder | | |
| Decoder Blocks | 300 | |
| Add Res. Blocks | 300 | |
| SCCB | | |
| Dilation Rate 5 | 25 | |
| Dilation Rate 11 | 25 | |
| Conv 1 | 6 | |
| Conv 2 | 6 | |

Table 1. Number of filters in convolution layers in each block

A sample input image and ground truth is shown in Figure 7.

### 3.2 Training

Each 6000x6000 image was cropped into 1024x1024 tiles with a 66% overlap along width and height. The last column and row of tiles was zero padded to create tiles of uniform size. We augmented the dataset by rotating each image by 90, 180 and 270 degrees to allow for better rotational uniformity in the representation of features. We normalized each DSM input tile using Equation 1. We normalized R-G-B-IR channels which are fed to the primary encoder as per Equation 2. We also used a Normalized Difference Vegetation Index (NDVI) channel derived from IR and red channels using Equation 3. Finally, we concatenated the channels grouped by their respective encoders and performed an average pooling with a pool size of 2x2 and stride 2 to down-sample the inputs to the encoders by a factor of 2 for the sake of computational efficiency. We used 10 percent of cropped tiles for validation during training. The tiles overlapping the validation set were removed from the training set. The output of the network was up-sampled by a factor of 2 to match size of the original input.

$$X_{norm} = \frac{X}{100} - 1 \qquad (1)$$

$$DSM_{norm} = (DSM - \mu(DSM))/35 \qquad (2)$$

$$NDVI = \frac{IR - RED}{IR + RED} \qquad (3)$$

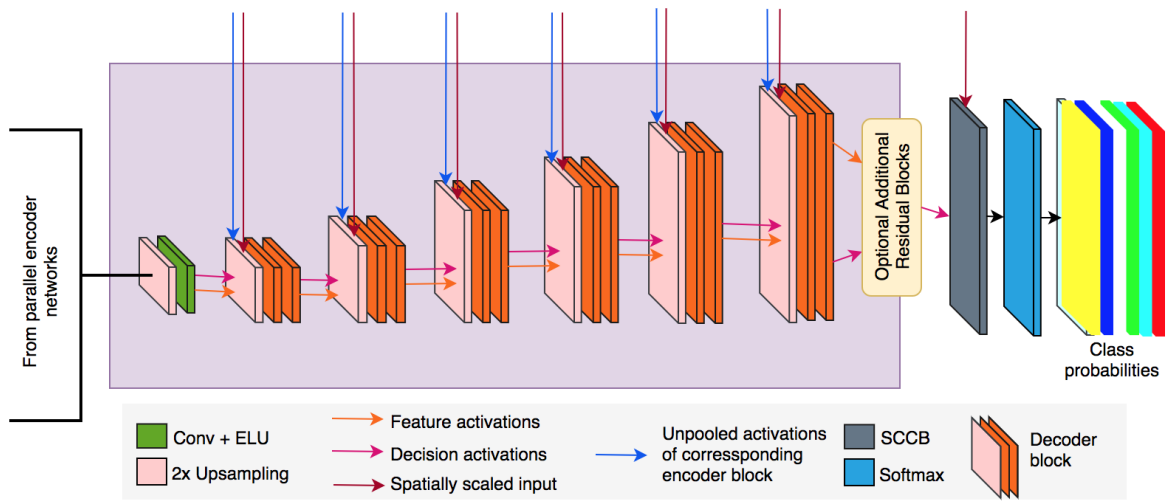

Figure 5. Decoder Connectivity

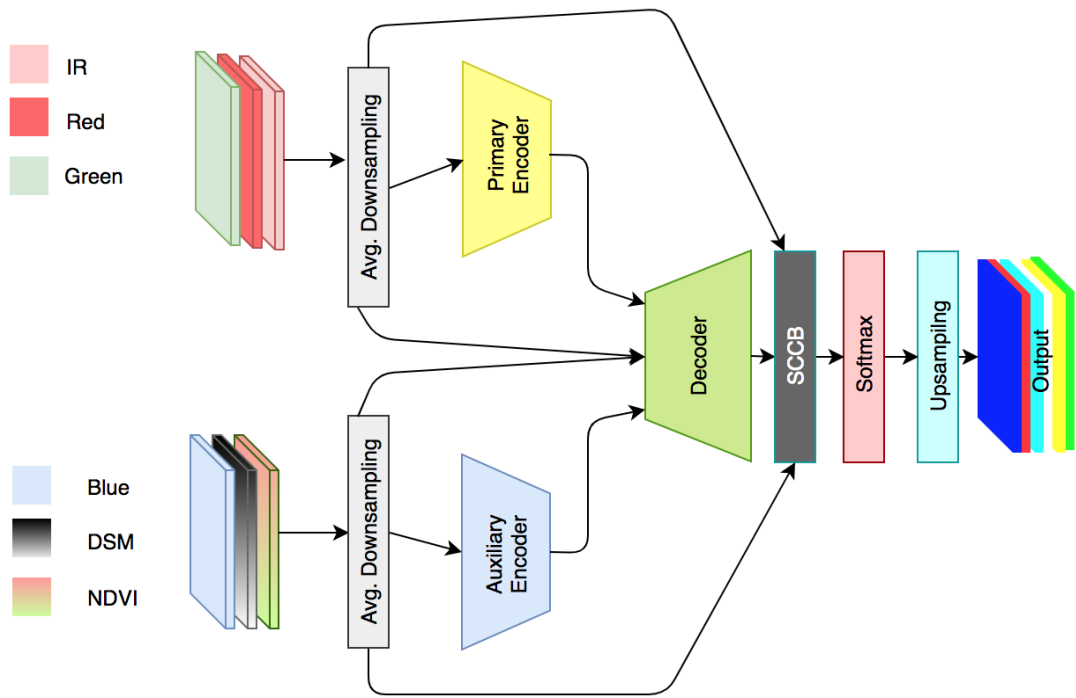

Figure 6. Overview of an instance of OrthoSeg used for Potsdam Benchmark

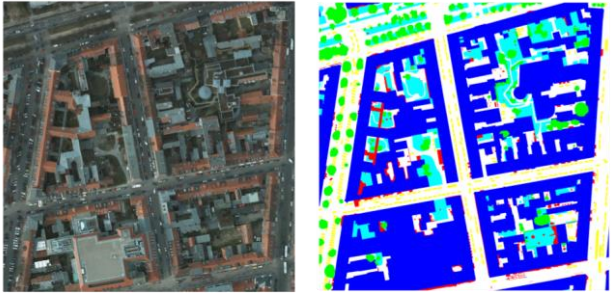

Figure 7. Image and ground truth sample

For training the network for the benchmark task, we used stochastic gradient descent (Kiefer and Wolfowitz, 1952) using nesterov accelerated gradient (p=0.99) (Sutskever et al., 2013). We used step decay learning rate schedule with initial learning rate of 0.0001 and reduced the learning rate by a factor of 10 when the validation loss plateaued for 25,000 iterations during training. The network was trained for 350,000 iterations. After first validation loss plateau, the momentum was increased to p=0.999. We refer to this phase as the fine tuning phase. A non-trainable activation scaling layer was used to scale inputs to first decoder block and the output from the last decoder by dividing the activations by 6 and 20 respectively. This reduces the chance of saturating the softmax layer during initial phases of training which can prevent the network from learning. We froze the first two encoding blocks of the primary encoder during the initial phase. We also employed depth-wise multiplicative gaussian noise during training for preventing over-fitting. We implemented this as a regularizing layer in our network, which is only active during training. This layer scales input activations of each filter independently by multiplying it with a scalar value n sampled from a gaussian distribution as per Equation 4 during each iteration of gradient descent. Here noiserate is a tuneable hyper-parameter. We term this noising method depth-wise multiplicative gaussian noise (DMGN) as it is a popular convention to stack filters along the tensor depth i.e. the last dimension. We reduced noiserate in the encoding blocks, decoding blocks and SCCB by multiplying it with a factor of 0.75, 0.375 and 0.25 respectively each time validation loss plateaued in the fine tuning phase.

$$n \sim N(1, \sqrt{\frac{noiserate}{1-noiserate}}) \quad (4)$$

DMGN has the similar effect as dropout (Srivastava et al., 2014) and it promotes network sparsity with the added advantage that it does not require rate dependent feature scaling during inference. In the encoder, we use DMGN after max pooling block for all the encoding blocks and additionally also between successive convolution layers in the fourth and subsequent encoding blocks. In the decoder block, we add noise to incoming feature activations and activations from the encoder and between convolution layers but not on the residual connections for decision activations. The noiserate used in encoder and decoder blocks for training the network for the Potsdam challenge is given in Figure 2. For layers in SCCB, we used noiserate value of 0.0625 for the output activations of dilated convolutions and did not noise the outputs of 1x1 convolution layers. We unfroze the second encoding block in the primary encoder after the first validation loss plateau in the fine tuning phase and unfroze the first encoding block after the subsequent plateau.

| Number of input feature maps | *noiserate* |
|---|---|
| ≤ 64 | 0.0625 |
| >64 and ≤128 | 0.125 |
| >128 and ≤256 | 0.1875 |
| >256 and ≤512 | 0.25 |
| SCCB(all) | 0.0625 |
| Additional Residual Blocks (all) | 0.0625 |

Table 2. *noiserate* used during training

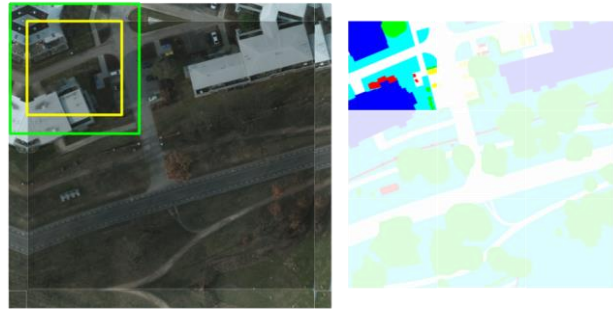

Figure 8. Overlap tile scheme for full image inference

The network was trained and tested using Keras (Keras Development Team, 2015) with Tensorflow (Tensorflow Development Team, 2015) backend on a single NVidia GeForce GTX-1080Ti GPU for approximately 82 hours.

### 3.3 Results and Discussion

We quantify the performance of OrthoSeg on the ISPRS Potsdam 2D semantic labelling challenge dataset using benchmark evaluation criteria described in (Gerke, 2015) on the 14 test images. Network performance compared to the previous best published results is summarized in Table 3. For running inference on the test data, we use the overlap-tile scheme (Ronneberger et al., 2015) to segment the entire high resolution image of 6000x6000 pixels by taking overlapping crops of size 1024x1024 pixels with stride of 256 pixels and extracting only the central region of 512x512 pixels for each tile. This allows to get rid of border artefacts and provides a seamless segmentation map since only the region for which full context is available is taken into consideration. For border pixels, we use symmetric padding of 256x256 in order to pad missing context, same as (Ronneberger et al., 2015). The method is illustrated in Figure 8. Sample output from the network and pixel-wise comparison with ground truth is shown in Figure 9.

### 4. CONCLUSION

OrthoSeg is an efficient architecture for semantic segmentation and is capable of achieving state-of-the-art results without the computational and conceptual complexity of using one-vs-all ensembles or explicit multi-scale context aggregation. Given the flexibility of OrthoSeg in terms of field-of-view and independent scaling of number of multimodal feature maps at different scales, it can serve as a useful architectural option for the purpose of semantic segmentation of multimodal overhead remote sensing imagery captured from aerial platforms and satellites.

| Method | Imp. Surf. | Building | Low veg. | Tree | Car | Overall Acc. |
|---|---|---|---|---|---|---|
| SVL – features + DSM + Boosting (**SVL 3**) | 84.0 | 89.8 | 72.9 | 59.0 | 69.8 | 77.2 |
| CNN + DSM + SVM (**GU**) | 87.1 | 94.7 | 77.1 | 73.9 | 81.2 | 82.9 |
| CNN + NDSM + Deconvolution (**UZ 1**) | 89.3 | 95.4 | 81.8 | 80.5 | 86.5 | 85.8 |
| CNN + DSM (**AZ 1**) | 91.4 | 96.1 | 86.1 | 86.6 | 93.3 | 89.2 |
| SegNet + NDSM (**RIT 2**) | 92.0 | 96.3 | 85.5 | 86.5 | 94.5 | 89.4 |
| FCN + DSM + RF + CRF (**DST 2**) | 91.8 | 95.9 | 86.3 | 87.7 | 89.2 | 89.7 |
| ResNet (**CASIA2**) | 93.3 | 97.0 | 87.7 | 88.4 | 96.2 | 91.1 |
| **Ours (ORTHOSEG)** | 94.8 | 97.7 | 88.7 | 88.4 | 95.7 | 92.5 |

Table 3. Benchmark performance comparison. All numbers in class columns are f1 scores and overall acc. is in percentage.

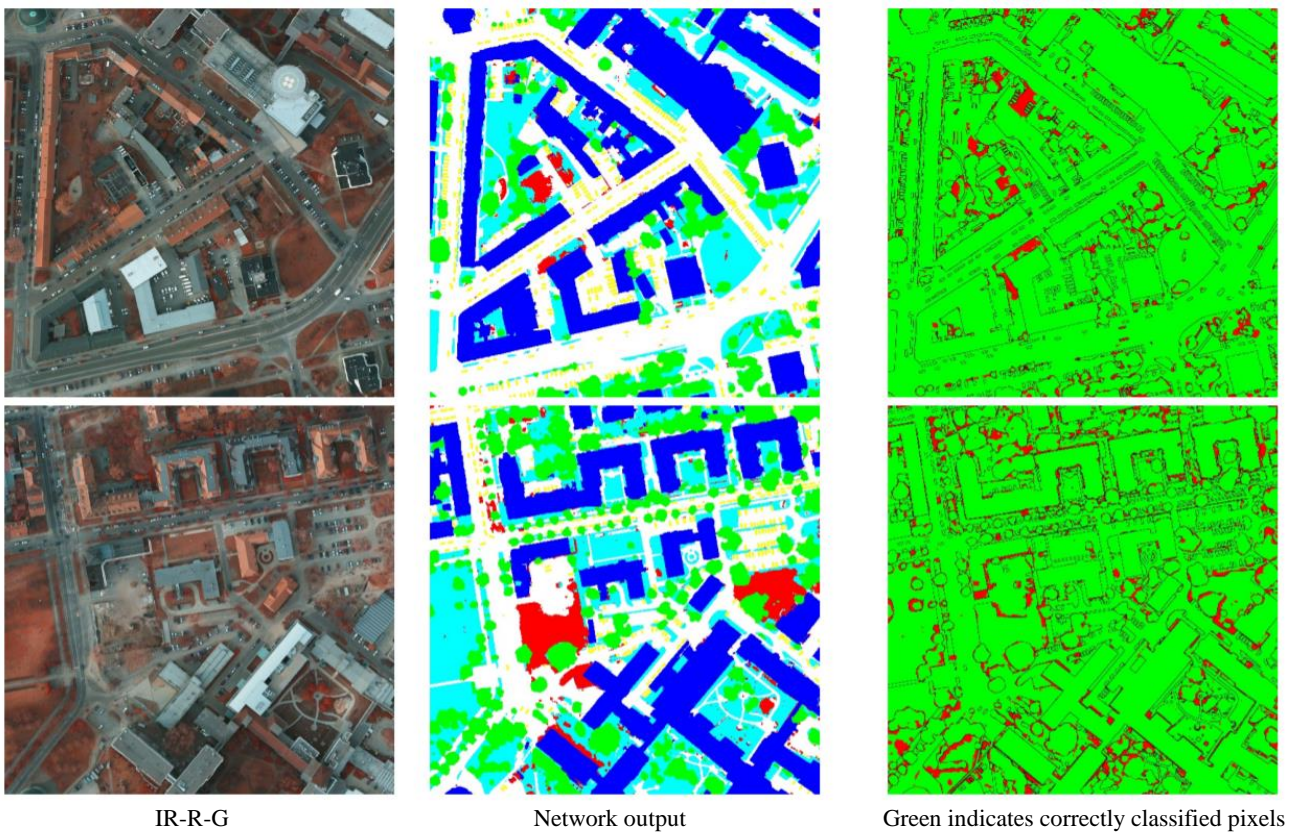

IR-R-G      Network output      Green indicates correctly classified pixels

Figure 9. Sample network output

## ACKNOWLEDGEMENTS

The authors would like to thank Shri. D.K. Das (Director), Dr. Rajkumar (Dy. Director, EPSA), Dr. Markand Oza and Dr. A.S. Rajawat at Space Applications Centre, Ahmedabad for their reviews, guidance and institutional support in carrying out this work.